\documentclass[a4paper,conference]{IEEEtran}

\newcommand{\linebreakand}{%
  \end{@IEEEauthorhalign}
  \hfill\mbox{}\par
  \mbox{}\hfill\begin{@IEEEauthorhalign}
}

\usepackage{graphicx}
\usepackage{subfig}
\usepackage{makecell}
\usepackage[table,xcdraw]{xcolor}
\usepackage{url}
\usepackage{array}
\usepackage{multirow}
\usepackage{comment}
\usepackage{hyperref}

\newcolumntype{P}[1]{>{\centering\arraybackslash}p{#1}}
\newcolumntype{M}[1]{>{\centering\arraybackslash}m{#1}}
\ifCLASSINFOpdf

\else

\fi

\begin{document}

\title{A Novel Disaster Image Dataset and Characteristics Analysis using Attention Model}

% Moinul islam Zaber, CSE, Dhaka University, zabermi@gmail.com 

\author{\IEEEauthorblockN{Fahim Faisal Niloy, 
Arif, Abu Bakar Siddik Nayem, Anis Sarker, Ovi Paul, \\ M. Ashraful Amin, Amin Ahsan Ali, Moinul Islam Zaber\IEEEauthorrefmark{2}, AKM Mahbubur Rahman}

\IEEEauthorblockA{AGenCy Lab, CSE, Independent University, Bangladesh; \IEEEauthorrefmark{2}CSE, Dhaka University, Bangladesh
}
niloy9542@gmail.com, 
[1611041, 1510190, 1521745, 1531144, aminmdashraful, aminali]@iub.edu.bd,\\ \IEEEauthorrefmark{2}zabermi@gmail.com, akmmrahman@iub.edu.bd}

\maketitle
% \IEEEauthorblockA{School of Electrical and\\Computer Engineering\\
% Georgia Institute of Technology\\
% Atlanta, Georgia 30332--0250\\
% Email: http://www.michaelshell.org/contact.html}
%\and
%\IEEEauthorblockN{Arif Hossain}
% \IEEEauthorblockA{Twentieth Century Fox\\
% Email: homer@thesimpsons.com}
%\and
%\IEEEauthorblockN{Abu Bakar Siddik Nayem}
% \IEEEauthorblockA{Starfleet Academy\\
% San Francisco, California 96678--2391\\
% Telephone: (800) 555--1212\\
% Fax: (888) 555--1212}
%\and
%\IEEEauthorblockN{Anis Sarker}
% \IEEEauthorblockA{School of Electrical and\\Computer Engineering\\
% Georgia Institute of Technology\\
% Atlanta, Georgia 30332--0250\\
% Email: http://www.michaelshell.org/contact.html}
%\and
%\IEEEauthorblockN{Ovi Paul}
% \IEEEauthorblockA{Twentieth Century Fox\\
% Email: homer@thesimpsons.com}
%\and
%\IEEEauthorblockN{M. Ashraful Amin}
% \IEEEauthorblockA{Starfleet Academy\\
% San Francisco, California 96678--2391\\
% Telephone: (800) 555--1212\\
% email: }
%\and
%\IEEEauthorblockN{Amin Ahsan Ali}
% \IEEEauthorblockA{Starfleet Academy\\
% San Francisco, California 96678--2391\\
% Telephone: (800) 555--1212\\
% email: }
%\and
%\IEEEauthorblockN{AKM Mahbubur Rahman}}
% \IEEEauthorblockA{Starfleet Academy\\
% San Francisco, California 96678--2391\\
% Telephone: (800) 555--1212\\

%\maketitle

\hypersetup{
    colorlinks=true,
    linkcolor=blue,
    filecolor=magenta,      
    urlcolor=cyan,
    pdftitle={Overleaf Example},
    pdfpagemode=FullScreen,
    }

\urlstyle{same}


\begin{abstract}
The advancement of deep learning technology has enabled us to develop systems that outperform any other classification technique. However, success of any empirical system depends on the quality and diversity of the data available to train the proposed system. In this research, we have carefully accumulated a relatively challenging dataset that contains images collected from various sources for three different disasters: fire, water and land. Besides this, we have also collected images for various damaged infrastructure due to natural or man made calamities and damaged human due to war or accidents. We have also accumulated image data for a class named non-damage that contains images with no such disaster or sign of damage in them. There are 13,720 manually annotated images in this dataset, each image is annotated by three individuals. We are also providing discriminating image class information annotated manually with bounding box for a set of 200 test images. Images are collected from different news portals, social media, and standard datasets made available by other researchers. A three layer attention model (TLAM) is trained and average five fold validation accuracy of 95.88\% is achieved. Moreover, on the 200 unseen test images this accuracy is 96.48\%. We also generate and compare attention maps for these test images to determine the characteristics of the trained attention model. Our dataset is available at \url{https://niloy193.github.io/Disaster-Dataset}
% We have also used class activation map (CAM) model to perform the same experiments. We got average accuracy of $96.23\%$ for five fold cross validation and accuracy of $95.48\%$ for the unseen test set.

\end{abstract}


\IEEEpeerreviewmaketitle

\section{Introduction}
%Necessity of Disaster Image Classification
%Due to the wide use of social media and online news papers, the news and images of occurrences of the disasters are very much available in real-time throughout the internet. 
%Bangladesh is one of the most vulnerable countries that suffer from huge climate change, as well as disasters. 
%Bangladesh’s flat topography, low-lying and climatic features are responsible for  
%Day after day, people are suffering from different kinds of disasters (e.g., Fire incidents in Chawk Bazar, Banani Fire incident) \cite{13}. These incidents which happened in March 2019 were featured by some newspapers such as the famous online news portal Dhaka Tribune [12], The Daily Star [13], etc.

\begin{figure}[t]

 \centering
  \includegraphics[height=6cm,width=0.5\textwidth]{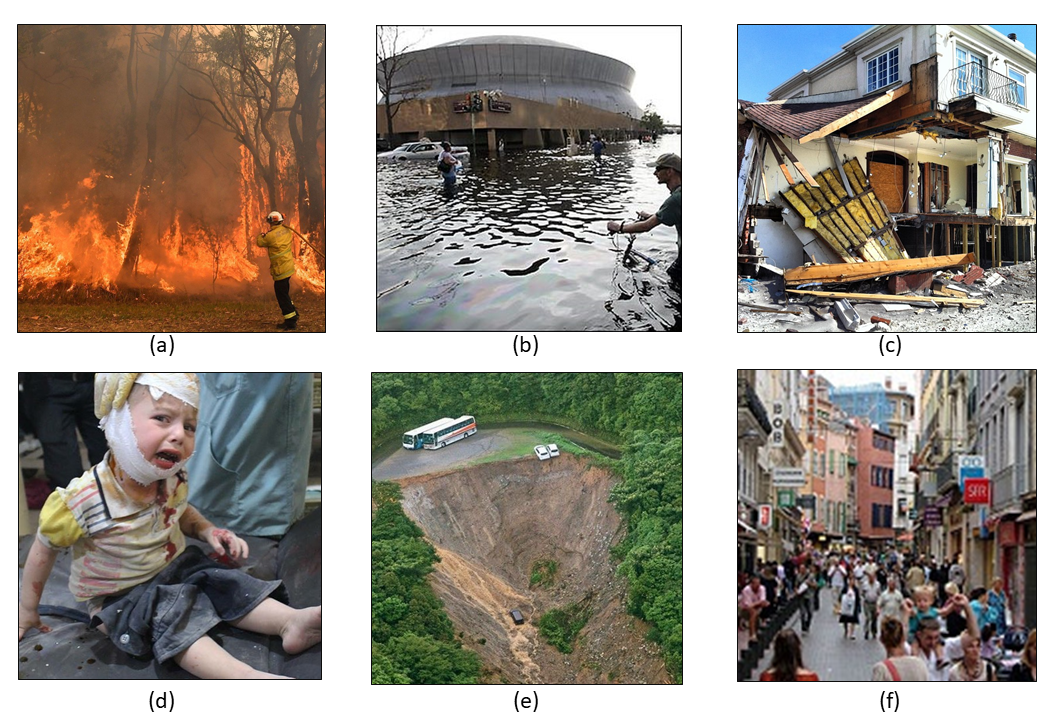}
 \caption{ (a) Fire Disaster, (b) Water Disaster, (c) Infrastructure Damage, (d) Human Damage, (e) Land Disaster and (f) Non-Damage  }
 \label{fig:imageIntro}
\end{figure}

In recent days, natural disasters, i.e., floods, cyclones, droughts, and earthquakes are becoming more common due to climate change, world-wide temperature rise, and pollution.  Moreover,  population density and socio-economic environments cause human-made disasters that include fire, building collapse, infrastructural damage, road accident, and armed war etc. Situations are getting worse in the developing countries that have very high density populations along with weak socio-economic structures. Generally, thousands are affected by  these disasters. Therefore, it is crucial in times of crisis that emergency response workers reach at the affected premises promptly to save human lives and prevent loss.  It would be great to have a system that would raise an alert and quantify the degree of damage of any disaster and inform the appropriate authorities based on an automated analysis of the images that are almost available in real-time on various social media. However, the state-of-the art deep learning techniques are not able to classify disaster types  from images due to lack of standard disaster datasets.  Existing disaster datasets have many limitations such as insufficient categories, imbalanced classes, wrong annotations etc. Therefore, in this paper, we propose an elaborated and standard dataset that has disaster images collected from google, twitter, facebook and other social media sites, online news portals and other standard datasets. Moreover, the proposed dataset also contains images of recent disasters: wild fires in Australia, flood in  India, forest fire in Amazon, and many more.  We have performed a number of experiments to show that the dataset can help build effective classifier models. Examples of different disaster images are shown in figure \ref{fig:imageIntro}.

% remove last three images not relevant here

%Brief description of existing disaster image datasets and image classifications
%Now we discuss existing disaster image classification techniques and the datasets used. 

A number of research has been conducted into classifying different kinds of disasters, both from deep learning and image processing domains in recent years. Different datasets have also been proposed for making learning process effective. Arif et al. \cite{OurGAIC2019paper} have collected  and experimented with South Asian Disaster (SAD) images that include disaster images from Bangladesh and other south asian countries. The authors here have observed that  the appearance of disaster images of south asia differ in various ways from western  disaster images. However, the main limitation of SAD dataset is that the images per class are too limited to be used to train deep learning models. 

\begin{table*}[!ht]
\setlength\tabcolsep{1.5pt}
\renewcommand{\arraystretch}{1}
\caption{Comparison of Several Existing Datasets}
\centering
\begin{tabular}{|c|c|c|c|c|c|}
\hline
Authors &
  Dataset &
  \begin{tabular}[c]{@{}c@{}}Number of\\ Classes\end{tabular} &
  \begin{tabular}[c]{@{}c@{}}Names of\\ the classes\end{tabular} &
  \begin{tabular}[c]{@{}c@{}}Number of\\ Images\end{tabular} &
  Comments \\ \hline
Rizk et al. \cite{rizk2019computationally} &
  Home-grown + Sun dataset &
  2 &
  \begin{tabular}[c]{@{}c@{}}Infrastructure and\\ Natural Disaster\end{tabular} &
  2344 &
  \begin{tabular}[c]{@{}c@{}}Contains only two\\ classes\end{tabular} \\ \hline
Giannakeris et al. \cite{panag2017people} &
  \begin{tabular}[c]{@{}c@{}}3F-emergency dataset\end{tabular} &
  2 &
  \begin{tabular}[c]{@{}c@{}}Fire and Flood \\ Disaster\end{tabular} &
  12000 &
  \begin{tabular}[c]{@{}c@{}}Contains only two classes,\\ Low diversity\end{tabular} \\ \hline
Muhammad et al. \cite{Khan2018Early} &
  \begin{tabular}[c]{@{}c@{}}Ko et al. \cite{ko2011modeling} + Verstock et al. \cite{ 2nddataset} + \\ Chino et al. \cite{bowfire} + Foggia et al. \cite{foggiadataset}\end{tabular} &
  2 &
  \begin{tabular}[c]{@{}c@{}}Fire Disaster and\\ Non-Damage\end{tabular} &
  68457 &
  \begin{tabular}[c]{@{}c@{}}Contains only two \\ classes\end{tabular} \\ \hline
Alam et al. \cite{firoj2017image4act} &
  Image4act &
  4 &
  \begin{tabular}[c]{@{}c@{}}Nepal Earthquake, Ecuador\\ Earthquake, Typhoon Ruby, \\ and Hurricane Matthew\end{tabular} &
  34562 &
  \begin{tabular}[c]{@{}c@{}}Natural disasters only,\\ Limited to narrow\\ geographical regions\end{tabular} \\ \hline
Arif et al. {[}1{]} &
  South Asia dataset (SAD) &
  6 &
  \begin{tabular}[c]{@{}c@{}}Fire Disaster, Flood Disaster,\\ Infrastructure, Nature Disaster,\\ Human Damage and Non Damage\end{tabular} &
  493 &
  \begin{tabular}[c]{@{}c@{}}The images per class are\\ very few\end{tabular} \\ \hline
Mouzannar et al. \cite{hussein2018multi} &
  \begin{tabular}[c]{@{}c@{}} UCI dataset \end{tabular} &
  6 &
  \begin{tabular}[c]{@{}c@{}}Fire Disaster, Flood Disaster,\\ Infrastructure, Nature Disaster,\\ Human Damage and Non Damage\end{tabular} &
  5880 &
  \begin{tabular}[c]{@{}c@{}}Non-damage class \\ contains irrelevant images, \\ Small number of images per class,\\ Low diversity, Dataset-bias\end{tabular} \\ \hline
  Niloy et al.&
  Proposed dataset \cite{ourdataset} &
  6 &
  \begin{tabular}[c]{@{}c@{}}Fire Disaster, Flood Disaster,\\ Infrastructure, Land Disaster,\\ Human Damage and Non Damage\end{tabular} &
  13720 &\begin{tabular}[c]{@{}c@{}} Several subcategories, \\ High diversity,\\Covers broad geographical regions,\\Natural and man-made disasters, \\Reduced bias \end{tabular} \\ \hline
\end{tabular}
\label{summary_ref_table}
\end{table*}

%Niloy, for following paragraph,  please provide the references in the IEEEICPR2020.bib file and use the reference here
%In paper \cite{kashif}, the authors have used Medieval\cite{Avgerinakis2017VisualAT} dataset and their own collected dataset for training deep convolutional neural network. The final dataset comprised of eight different disasters: cyclone, drought, earthquake,  ﬂood, landslide, thunderstorm, snowstorm, and wildﬁre. They extracted the global feature vector and trained different classifiers (eg: Support Vector Machine -SVM, K-Nearest Neighbors-KNN, etc); SVM yielded the best result. Limitation of this work is the only use of one type of disasters: natural disaster. 

In the paper \cite{Khan2018Early}, the authors extracted a total of 68457 images from  dataset \cite{ 2nddataset} and Chino dataset \cite{bowfire}. From Foggia dataset \cite{foggiadataset}, they collected video frames. They used an architecture similar to Alexnet \cite{alexnet} and achieved state-of-the-art results. %However, they have experimented with fire disaster and non-damage classes only. 
%The dataset only contains 226 images, where 119 images contain fire and rest of 107 are fire-like images containing fire-like lights, sunsets,  and sunlight coming through windows  \textbf{[? limitations done]}. 

% In \cite{firoj2017image4act} authors created dataset from twitter, which were subsequently annotated using crowd sourcing. [limitations, ? How many classes]. They evaluated their architecture in real time in the event of cyclone Debbie.

%For severity estimation, as no such dataset was found, t

In the work \cite{panag2017people}, the authors classified, localized, and estimated severity of different disasters. They used MediaEval \cite{Avgerinakis2017VisualAT} dataset for classification and Bow fire dataset \cite{bowfire} for localization. They developed their own dataset named 3F emergency dataset that consisted of flood and fire pictures taken from flicker. Their classification algorithm surpassed other participants in accuracy metric of MediaEval challenge. However, their datasets suffer from inadequate disaster events. They have performed experiments with only flood and fire classes. Rizk et al. \cite{rizk2019computationally} proposed a multi-modal two-stage framework that relies on computationally inexpensive visual and semantic features to analyze Twitter data. In this paper, two datasets were used: Home-grown and Sun dataset. Home-grown dataset comprises of Twitter images which only covers damaged infrastructure and natural disaster. Sun dataset was made from several search engines and it also contains infrastructure and natural disaster. So, these datasets contain two categories only. Moreover,  the size of their dataset is very small: only 2344 images. 
% The above part is repeated 

%Deep Convolutional Neural Network (DCNN), DBpedia Spotlight \cite{} and combMAX \cite{} were implemented to tackle DIRSM.
In \cite{Avgerinakis2017VisualAT}, the researchers present the algorithms that the team deployed to tackle disaster recognition tasks. They made two flood disaster dataset: one of them was made from social media image and the other one from satellite images. GoogleNet architecture was used to train on the images. A major limitation of their dataset is that it only contains flood disaster category. Alam et al. \cite{firoj2017image4act} proposed  an image filtering module that employs deep neural networks and perceptual hashing techniques to determine whether a newly-arrived image is relevant for a given disaster response context. To train the relevancy filter, 3,518 images were randomly selected from the severe and mild categories. The authors have collected four types of natural disasters: Nepal Earthquake, Ecuador Earthquake, Typhoon Ruby, and Hurricane Matthew. No other regional disaster images are present in the dataset.
%/Arif: Describe the dataset briefly % also provide one or two limitations of 

% \begin{figure}[!t]
% \label{fig:attentioncam}
%  \centering
%   \includegraphics[height=5cm,width=0.52\textwidth]{6cams.png}
%   \caption{    }
% \end{figure}
%Repetitive paragraph
%Giannakeris et al. \cite{panag2017people} presented a novel warning system framework for detecting people and vehicles in danger. The proposed framework provides a near real-time localization solution for detecting and scoring severity and safety levels of people and vehicles in flood and fire images. They chose to fine-tune the pre-trained parameters of the VGG-16 on Places365 dataset to leverage useful distinctions between various visual clues that relate to generic scenery images. Their Initial 3F-emergency dataset is composed of 6K images from Flickr. In this paper they have made their own dataset by collecting images from Flickr and other social media sources and calls it Fire-Flood Flickr (3F) emergency dataset. This dataset only contains two category of disaster those are fire and flood; %which is the  limitation of this dataset. 
%/Arif: Describe the dataset briefly % also provide one or two limitations of this dataset

%/Arif: Describe the dataset briefly % also provide one or two limitations of this dataset

%The inception convolutional neural network (CNN) model was adopted, which was pre-trained on ImageNet to process images, and for words, they used a pre-trained word embedding model to process the texts.

\begin{figure}[t]

 \centering
  \includegraphics[height=5cm,width=0.50\textwidth]{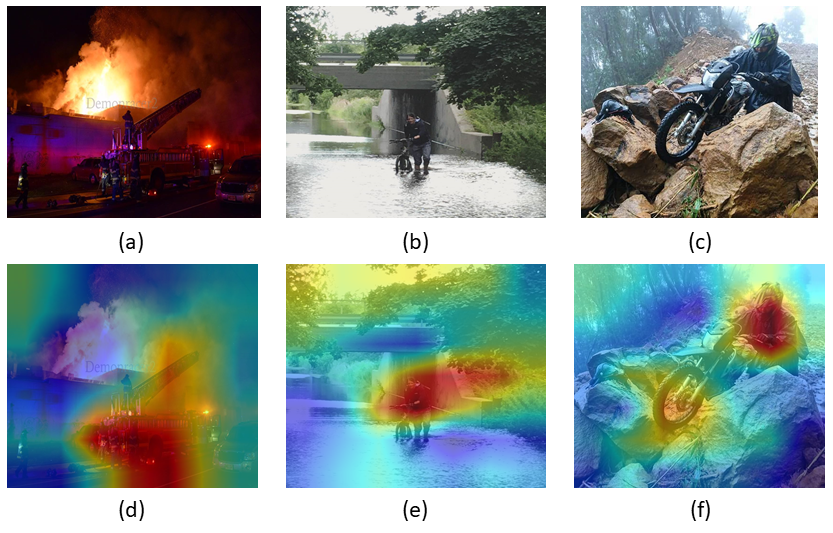}
  \caption{ First row contains input images from different disaster classes, second row contains corresponding attention heatmaps;
  (a) Fire Disaster (b) Water Disaster (c) Infrastructure Damage (d) Misplaced attention (vehicle); attention should be focused on fire region (e) Misplaced attention (vehicle and human); attention should be focused on water region (f) Misplaced attention (human, motorcycle); attention should be provided to the damaged infrastructure region}
  \label{fig:attentionIntro}
\end{figure}

Mouzannar et al. \cite{hussein2018multi} proposed a multimodal deep learning framework to identify damage related information from social media posts. This framework combines multiple pretrained unimodal convolutional neural networks that extract features from raw texts and images separately. The framework was evaluated on a homegrown labeled dataset that contains images collected from  social media posts. Their dataset (UCI dataset) contains six categories: fire, flood, natural disaster, infrastructure damage, and non-disaster. Though this dataset contains images for various disaster events, their non-disaster class contains lots of irrelevant images, e.g, images of foods, products, jewelries etc. Keeping these images into non-disaster class might result in good  overall classification performance but learned models would not distinguish between damaged  and undamaged infrastructures. Another limitation is, the deep models trained on UCI dataset do not show well attention localization capability, because the dataset is not much diverse. For example, we observed that the models trained on UCI dataset use fire-trucks in the disaster image as a proxy to classify fire disaster event, which is not expected. Some examples of misplaced attentions are shown in figure \ref{fig:attentionIntro}. 

%Therefore, using the dataset for training classifiers might produce biased and incorrect model.

%the limitation of this dataset is it only contain 5879 images and their  

%Arif: Describe the dataset briefly % also provide one or two limitations of this dataset
% In \cite{damageassess} damage severity of disasters was detected.  The dataset consisted of annotated pictures extracted from social media using AIDR platform in the event of disasters. Also a handful of images was taken from web after disaster which were subsequently annotated using Crowdflower.

In the paper \cite{damageassess}, images posted on social media platforms during natural disasters are analyzed  to determine the severity of damage caused by the disasters. The authors collected images of different disasters: Typhoon Ruby, Hurricane Matthew, Nepal Earthquake from internet.  The authors also used google search to collect images like damaged building, damaged bridge, damaged road etc. The limitation of the dataset is that it only contains damaged infrastructure images.

%Arif: Describe the dataset briefly % also provide one or two limitations of this dataset

%They ran tests on their training sets, and the highest Precisionuracy they achieved was when they combined their Google, Ruby and Matthew datasets.
%In this research paper, we propose a deep learning-based method to automate the effective extraction of information from social media posts to direct relief resources efficiently. 

%Limitations of existing Disaster datasets and classfication techniques
Summary statistics of the datasets are shown in table \ref{summary_ref_table}. From the datasets mentioned above, it is easy to notice that  a benchmark dataset for disaster classification is yet to be published. In most of the literature, a scarcity of benchmark dataset is clearly seen. In summary, the limitations are:
\begin{itemize}
\item	The datasets do not cover broad regions. A comprehensive dataset should have disaster images from most major regions of the world.

\item After training deep learning architectures on several existing disaster datasets, we observe that the classifiers show poor attention localization capability, because the datasets are not diverse enough. That is why the classification accuracy deteriorates. %This is discussed more in the experimental section.

% After training deep learning architectures on several existing disaster datasets (ex: UCI), it is found most often thealgorithms do not focus it’s attention (Fig-2) on disasterspecific  items  (ie:  fire,  water  etc.),  rather  it  infers  thedisaster  class  from  non  disaster  specific  items.  That  iswhy the classification accuracy deteriorates.

\item Most of the datasets do not contain images having enough challenging scenarios. As a result, the algorithms are prone to misclassification when exposed to semantically similar images. For example, it is common for architectures trained on fire disaster images to misclassify images with high brightness and reddish hue to be a fire disaster event.

\item A diverse dataset having good volume of disaster images with wide number of categories and subcategories is yet absent.

\end{itemize}

%How our dataset address above problems
To overcome the above limitations, we propose a novel dataset where we have collected images for a number of disaster events that include both natural and non-natural(man-made) disasters from different geographical regions. Also we have carefully hand-picked and annotated several test images which are used in separate attention models to show the efficacy of our proposed dataset. Most specifically our contributions are:

\begin{itemize}
\item A novel disaster dataset with 6 disaster categories and 10 subcategories, consisting of a total 13720 images. The detailed statistics of our proposed dataset is shown in table \ref{tab:dataset_statistics}

\item Bounding box annotated images for 200 test images. These are used in attention verification to show improved attention localization capability of classifiers trained on our dataset.

\item Detailed characteristics analysis using attention models. We use CAM\cite{zhou2016learning}, TLAM\cite{jetley2018learn} to show deep learning classifiers trained with our dataset yield better results compared to existing datasets.

\end{itemize}

\

%For better model development, standard dataset is a prerequisite \cite{ashraful2015overview}.
\section{Proposed Dataset Description}
%introductory description, Our goal and objectives
 A well described and diversified dataset is needed for deep learning systems to perform well in classification. Therefore the objective of this paper is to provide a well defined and diversified dataset for disaster classification. As most of the existing datasets do not contain disaster images from  major regions, our purpose is to create a dataset which contains images from all major regions (i.e. western, asian, tropical regions etc.). Existing datasets do not have well organized sub categories. Therefore, another of our objective is to create a dataset that contains well organized sub categories. %that would help researchers who want to classify disasters more specifically.
% \usepackage{multirow}

Moreoever, a good reason of deep learning networks' poor performance in classifying most existing disaster datasets is that the networks do not focus their attention on disaster related items (e.g. smoke, water etc.). Generally, disaster images have many subjects involved. For example, a cat image may have a cat which is the subject and simple background. A dog image may also have the same scenario. So it is easier to differentiate between a dog and a cat. However, disaster images may have wide range of semantics involved, i.e. there can be images with damaged buildings, crowd of humans, cluttered backgrounds etc. Consequently, it becomes tough for deep learning networks to focus on the correct region of interest. While training with UCI images, we have observed that the classifier learns the fire trucks as features for fire disaster since large number of fire images contain fire trucks. These types of misplaced attention might produce inappropriate features that would result in poor classification performance. For this reason, we have carefully collected thousands of images having wide range of varieties so that the network is forced to learn to focus it's attention on disaster related items. This also helps in improving classification performance which is shown in the experiment section. Paying attention to the appropriate regions also confirms the quality of classifier models.
% \usepackage{multirow}

\begin{figure}[t]

  \centering
  \subfloat[Urban Fire]{\includegraphics[width=4cm,height=3cm]{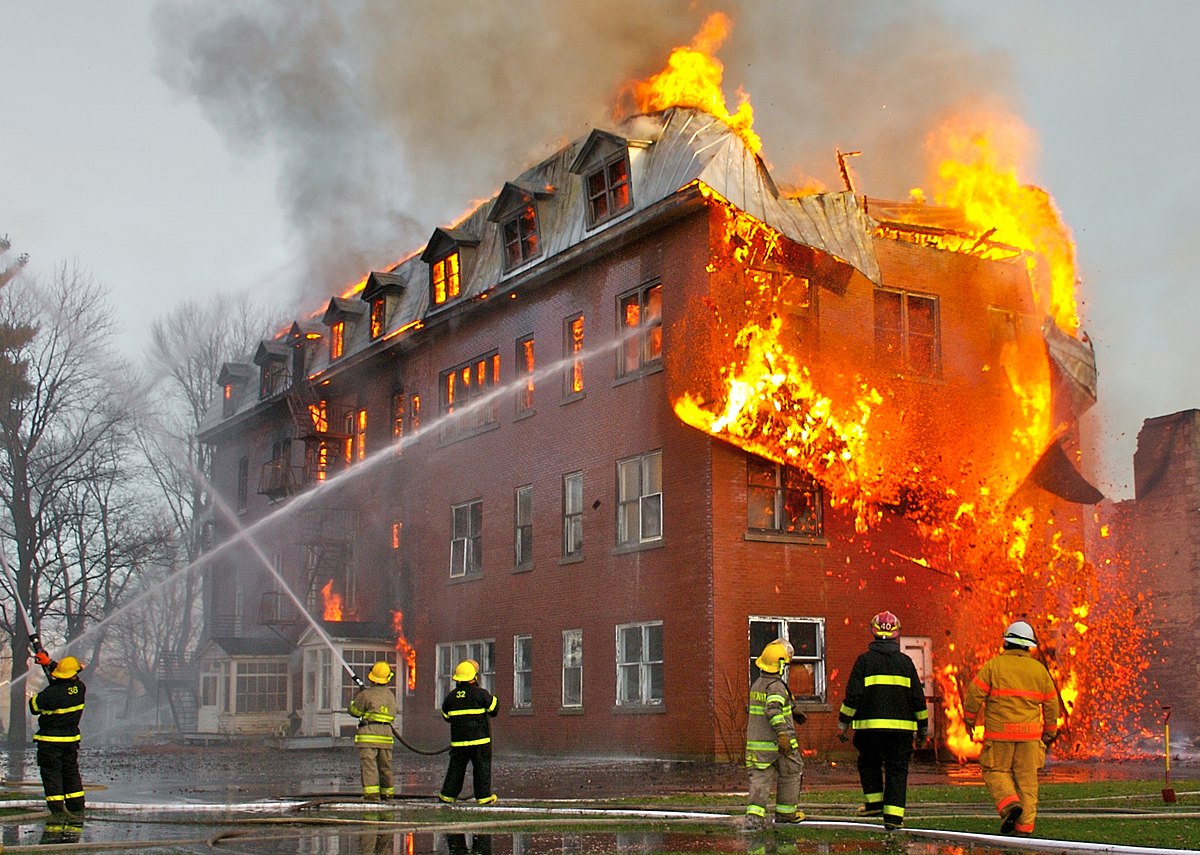}
  \label{fig:f1}} 
  \hfill
  \subfloat[Wild Fire]{\includegraphics[width=4cm,height=3cm]{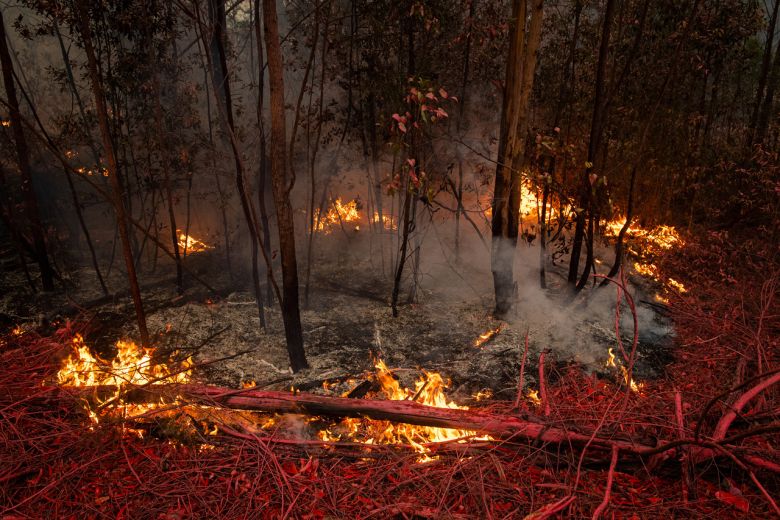}
  \label{fig:f2}}
  \caption{Subcategories of Fire Disaster}
  \label{urbanvswild}
\end{figure}

We have collected images for three types of disasters: Fire Disaster, Water Disaster, Land Disaster. Additionally, there are two damage related classes: Damaged Infrastructure, Human Damage. The sixth class is Non-Damage where normal images with various infrastructure, natural scene, forest, beach are grouped together. We have also added several sub categories: Urban fire and Wild fire in Fire Disaster category, Landslide and Drought in Land Disaster category etc. Figure \ref{urbanvswild} shows example images from fire subcategory. Moreover, we have created four subcategories for Non Damage class: Human, Building and Street, Wildlife Forest and Sea. The Non-Damage images are limited to four categories because we wanted to put a 'negative set' for each disaster category, e.g, the non-damage human sub-category can be considered as a negative set for “Human Damage” category. The same way, non-damage buildings and streets sub category is negative for “Damaged Infrastructure” or “Urban Fire”; non-damage sea is negative for “Water disaster” and non-damage forest can be considered a negative set for both "Land Disaster" and "Wild fire". This is also the reason why the Non-Damage category contains the most number of images. The proposed dataset is made publicly  available here \cite{ourdataset}.

% \usepackage{multirow}
\begin{table}[!ht]
\caption{Proposed Dataset Summary}
\label{tab:dataset_statistics}
\centering
\begin{tabular}{|c|c|c|c|c|}
\hline
Category                       & Sub-Category                                                   & Train & Test                & Total                 \\ \hline
\multirow{2}{*}{\begin{tabular}[c]{@{}c@{}}Damaged\\ Infrastructure\end{tabular}} & Infrastructure & 1418 & \multirow{2}{*}{34} & \multirow{2}{*}{1488} \\ \cline{2-3}
                               & Earthquake                                                     & 36    &                     &                       \\ \hline
\multirow{2}{*}{Fire Disaster} & Urban Fire                                                     & 419   & \multirow{2}{*}{33} & \multirow{2}{*}{966}  \\ \cline{2-3}
                               & Wild Fire                                                      & 514   &                     &                       \\ \hline
Human Damage                   & \multicolumn{1}{l|}{}                                          & 240   & 32                  & 272                   \\ \hline
Water Disaster                 & \multicolumn{1}{l|}{}                                          & 1035  & 33                  & 1068                  \\ \hline
\multirow{2}{*}{Land Disaster} & Land Slide                                                     & 420   & \multirow{2}{*}{33} & \multirow{2}{*}{654}  \\ \cline{2-3}
                               & Drought                                                        & 201   &                     &                       \\ \hline
\multirow{4}{*}{Non Damage}    & Human                                                          & 120   & \multirow{4}{*}{35} & \multirow{4}{*}{9272} \\ \cline{2-3}
                               & \begin{tabular}[c]{@{}c@{}}Building \\ and Street\end{tabular} & 4572  &                     &                       \\ \cline{2-3}
                               & Wildlife                                                       & 2271  &                     &                       \\ \cline{2-3}
                               & Sea                                                            & 2274  &                     &                       \\ \hline
Total Images                   & \multicolumn{1}{l|}{}                                          & 13520 & 200                 & 13720                 \\ \hline
\end{tabular}
\end{table}

\subsection{Disaster Image Collection}
% Arif
%Sources, How we collect, 
%Images from another dataset  Disaster\_1\cite{disaster1} are collected that contains building-fire, earthquake, flood, forest-fire, and non-disaster. 
%\cite{aussiebushfirexinhua1}\cite{aussiebushfirecnbc}
%\cite{californiawildfirenytimes}\cite{californiawildfirenbc}
%\cite{braziltime}

We have collected disaster images from different number of sources. Normal buildings, street, forest, and sea images are collected from google and have been put in non disaster category. A large number of different disaster images such as fire, earthquake, tsunami, landslide and flood have been collected from google and popular social media sites such as facebook, twitter etc. We have focused on notable recent disasters like Kerala floods from South India, Japan's tsunami for water disaster, Australian bushfires, California wildfires, Brazil Amazon rain forest wildfires, Hong Kong protests police violence, and so on for fire disaster.
Moreover, we have collected a number of disaster images by scrapping different news portals. Some notable examples are: California wildfires \cite{californiawildfireesquire}, Brazil wildfires \cite{brazilcbsnews} and many more. 
In Social media platforms, \emph{hashtag} categorizes similar type of posts or images. We have used \emph{hashtags} to collect image data for different kind of disasters from facebook and twitter. To collect images for human damage, we have focused on Syrian civil war, Yemeni civil war etc. Additionally, we have gathered some of the damaged infrastructure images from news portals \cite{portal:yemenwar}, facebook, twitter etc. We have taken several building fire, forest fire, damaged infrastructure, tsunami etc. images from SAD \cite{OurGAIC2019paper}. Also we have collected a lot of non disaster images from \cite{intelimage}. 
%\cite{aussiebushfirexinhua}
%Some examples are: Brazil Amazon fires, Land slides , tsunami, Australian bushfires , California wildfires etc.

After collecting images with above mentioned process, we gathered almost 16000 images in total. However, we had to discard few number of images which were very low resolution, had embedded texts etc. A wide number of images were later discarded in the course of annotation. We finally got a total of 13720 images. The image shapes of our dataset are diverse. During training, the images were resized to $224\times224$. The class statistics of the collected images are shown in table \ref{dataset_statistics}.

%The following is a tree diagram of the categories:

\subsection{Annotation of Disaster Image Categories} 
\label{subsec:assessment}
After collecting and cleaning up, the category for each image is determined by our well trained annotators. For this purpose, three human annotators have been trained beforehand on the image classification ideas and methodologies. They learned about different disaster classes and sub-classes. Also, they have gone through different video and image sources to understand the impact of different disasters. Each image from the collected set has been annotated by  three annotators separately without any knowledge about the annotation of others. 
%which is very common in the images collected from social-media sites.
%The annotators each created a spreadsheet which contains image names and the categories as columns. They later put a value of 1 to the corresponding class after inspecting each image separately. 
%Simultaneously, data cleaning is also done by them, which is crucial to get rid of redundant and unnecessary images. Images that contain multiple types of disasters that are not correlated (such as, fire and water disasters cannot happen at the same place) are discarded  as they are often the result of photo-collage rather than a single image. Photo-collages also impose other difficulties as often there are before after shots of the said disaster image is the subject of the collage. 

% Each image in the dataset contains the corresponding disaster object at such a scale and fashion that the category is determinable by human without being confused with other disaster or non-disaster images.
%

 For Water Disaster category, the defining characteristic is excessive amount of water present in undesirable places i.e, fields, roads, establishments that are fully or partially submerged in water due to floods, tsunami etc. Thus, these images are kept in a unified Water Disaster category. Similarly, the images with landslides are kept under Land Slide sub-category. The land images with drought are kept in the Drought subcategory. The Urban Fire images tend to have buildings, cars, traffic, and other types of infrastructures with fire whereas the wildfire images normally have trees and other types of greeneries, grasslands and often animals. Damaged Infrastructure category has images where there are broken remnants of buildings or concrete infrastructure, vehicles etc. The structural damages caused by earthquakes are kept under Earthquake subcategory. Human damage category consists of bloody, wounded, burned, and gory pictures due to war or accidents. Bandages and stretchers are also present in some of the images in the human damage category. 
%Additionally,  it is determinable from the images that there are natural harmony is missing from those objects due to some disaster very recently.

During these class label annotation tasks,  we have kept all three annotators' labels into account. If all three annotations differ, the image is discarded. If two annotations coincide, the image is put into that category. After this filtration process we have ended up with 13720 images in total.

%typical and common cat/dog example has been used to make them aware of what kind of classification we are doing as there exists many type of annotations for various tasks such as 2D/3D bounding box, polygon, cuboid, semantic segmentation, land-marking etc. Since our data is not as simple as a cat/dog image dataset by nature,
%How we assess the disaster category,  criteria behind your decisions

\subsection{Creating the training and test sets}
For our experiment, we have merged the sub classes of each parent class. Therefore, our training classes are Fire Disaster, Human Damage, Water Disaster, Land Disaster, Damaged Infrastructure, and Non-Damage. We have carefully handpicked two hundred challenging images and used that as test set. We have tried our best to ensure that these test images reflect the diversity of our dataset.% The number of images used in training and test are shown in Table \ref{tab:imagenumber}.

% We have made our training set with all the images except 200 images, that have been kept apart as unseen test data.

%Fine grained classification is a challenging domain on it's own. We may tackle the fine grained classification of the disaster dataset in future

Our test set consists of images from each parent class. Our primary focus is to keep as much variety as possible in the selected test images. Hence, we have included aerial images, landscapes, low light images, scenarios with many subjects in our test set. We have also put images that would be challenging for the network to classify, like non disaster images with red hue which resemble fire disaster image, sea-beach images which resemble water disaster etc. For each parent class that has sub-classes, we pulled images from each of the sub-classes. We didn't use any random method to collect the test images, rather the selection procedure was carefully performed by human selectors following the above guidelines.

% \begin{table}[!ht]
% \caption {Number of images used for training and test process} \label{tab:imagenumber}
% \begin{center}

% % For LaTeX tables use
% \begin{tabular}{lll}
% \hline\noalign{\smallskip}
% Process & Training & Test \\
% \noalign{\smallskip}\hline\noalign{\smallskip}
%     Number of images & 13238 & 200  \\ 
% \noalign{\smallskip}\hline
% \end{tabular}
% \end{center}
% \end{table}

% \begin{figure}[!tbp]
% \centering
% \begin{tabular}{lcc}
% \subfloat[]{\includegraphics[width=4cm,height=3cm]{infra.png}}&
% \subfloat[]{\includegraphics[width=4cm,height=3cm]{firedmg.png}}\\
% \subfloat[]{\includegraphics[width=4cm,height=3cm]{gt_infra.png}}&
% \subfloat[]{\includegraphics[width=4cm,height=3cm]{gt_firedmg.png}}\\
% \end{tabular}

% \caption{Infrastructure(a), fire(b) damage and their repective bounding box attention annotations(c), (d) }
% \label{fig:fig1} % I can do without the label too
% \end{figure}
%We complied with the rules and guidelines mentioned there that the research community has long been using to annotate images in order to create newer datasets.
% , meaning that the bounding box in an image contains non disaster class pixels as less as possible.

\section{Dataset Characteristics Analysis}

\subsection{Diversity Analysis}
One of the key characteristics of our proposed dataset is diversity. But, it is difficult to devise a measure that quantifies diversity. However, to tackle this problem we follow the procedures mentioned in \cite{deng2009imagenet}. We compute the average image of each class and measure lossless JPG file size which reflects the amount of information in an image. A diverse image class will result in a blurrier average image, consequently the JPG file size wil be smaller. On the other hand, a less diverse image class will result in a more structured, sharper average image with a greater JPG file size.

\subsection{Performance and Attention Analysis}
We design our experiments to show the efficacy of our dataset in training deep learning models. Moreover, the experiments are performed to show how the classification models put their attention on particular parts of the images for predicting class label. 
% \subsubsection{Our Experiment Objectives:} 
Our experimental objectives are:
\begin{itemize}
    
    \item To measure the quality of the training set for building classifiers. For this purpose, we use five fold cross validation to show the generalization ability across different folds of the dataset.
    
    \item To measure the performance of classifiers on unseen data.
    
    % \item To see how training dataset helps improve attention localization capability of classifiers. To do so, we perform experiments to find out the regions of input images where different models pay their attentions.
    
    \item Our final objective is to quantify the attention localization capability of classifiers trained on our dataset.
    % We also compare the attention capability of the classifiers are trained on state of the art UCI dataset. 
    
\end{itemize}

\subsubsection{Classifier Model Description}
We have used  VGG-16 as our classification network. The weights of the network are initialized with weights pre-trained on ImageNet data. The input image size is $224\times224$. To depict the region of input image where the network is focusing it's attention, two separate attention modules \cite{zhou2016learning}, \cite{jetley2018learn} are used:

\textbf{Class Activation Map (CAM):}
Upon the VGG-16 network, Global Average Pooling (GAP) is used. The outputs of the GAP layer goes to six class softmax layer. To get the attention heatmap the weights of the dominant class is later dot multiplied with the last convolutional layer of VGG-16, which is then upsampled to the input image size.

\textbf{Three Layer Attention Map (TLAM):}
The TLAM\cite{jetley2018learn} demonstrates how soft trainable attention can improve image classification performance and highlight key parts of images. We use a VGG-16 network where local feature maps $L_i (i=1,2,3)$ are taken from the last three maxpool layers. Then global feature map G is taken after the last conv layer. A parameterized compatibility score is calculated from which attention weights are found.  Finally, all three attention maps are concatenated and passed through a fully connected layer to get the final prediction. 
% \begin{equation}c_{i}^{s}=\left\langle\boldsymbol{u}, \ell_{i}^{s}+\boldsymbol{g}\right\rangle, i \in\{1 \cdots n\}\end{equation}
% The shape of G and $L_i$ are matched using a projector that stretches the number of channels of $L_i$ where it is lower than G.
%Using the attention weighted combinations of  $L_i$, final output of each attention layer is calculated.

\subsubsection{Five-fold Cross Validation}
%How you did cross validation
% Non-overlapping folds
% Randomly selected without replacement to make sure that an image has been selected only once in a fold.
%The distribution of the training images 
To evaluate the uniformity of the distribution of our dataset, we have performed five fold cross validation. In each validation process, we have selected $80\%$ images from each class for training and rest $20\%$ images for testing. We make sure that no images from this $20\%$ set are selected for testing in  other validation process. That means: the training and testing sets are always non-overlapping in each validation iteration.

\subsubsection{Testing}
%How you train
%how you test
%Training parameters
%Test parameters
%Batch size during train
%
We use both CAM and TLAM architectures for training and then test on the 200 unseen test data that we created. We have reported the classification result for both our dataset and UCI dataset. 

We have used a batchsize of 64 and 32 respectively for CAM and TLAM. We use a small learning rate (0.0001) to make sure effective fine tuning occurs. We have used Adam optimizer and weighted cross entropy loss during the training.

\subsubsection{Human assessment of visual  attention} 
\begin{figure}[t]

  \centering
  \subfloat[Fire Disaster]{\includegraphics[width=4cm,height=3cm]{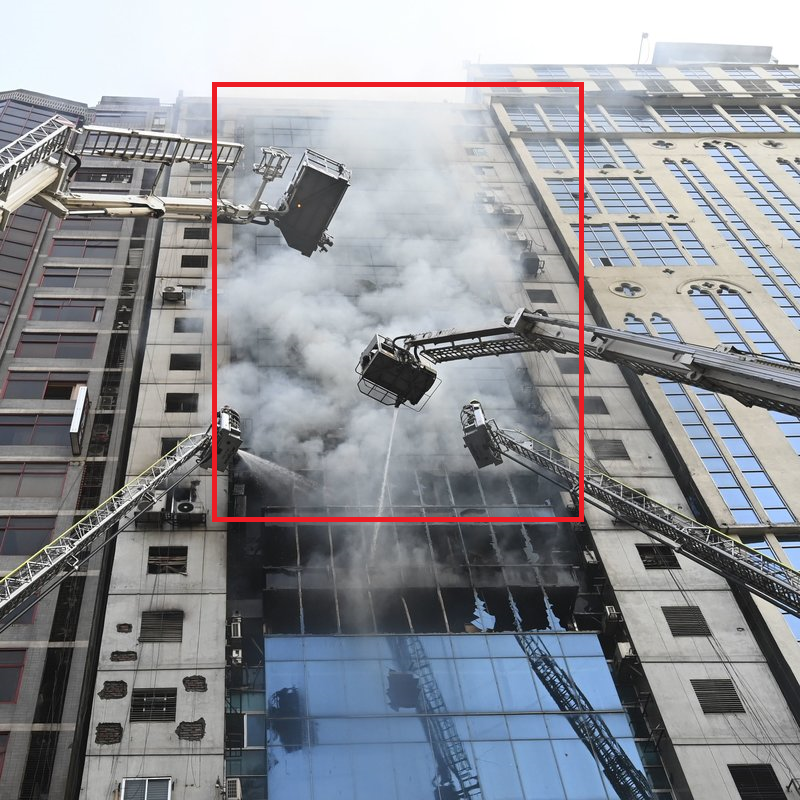}
  \label{fig:f1}} 
  \hfill
  \subfloat[Damaged Infrastructure]{\includegraphics[width=4cm,height=3cm]{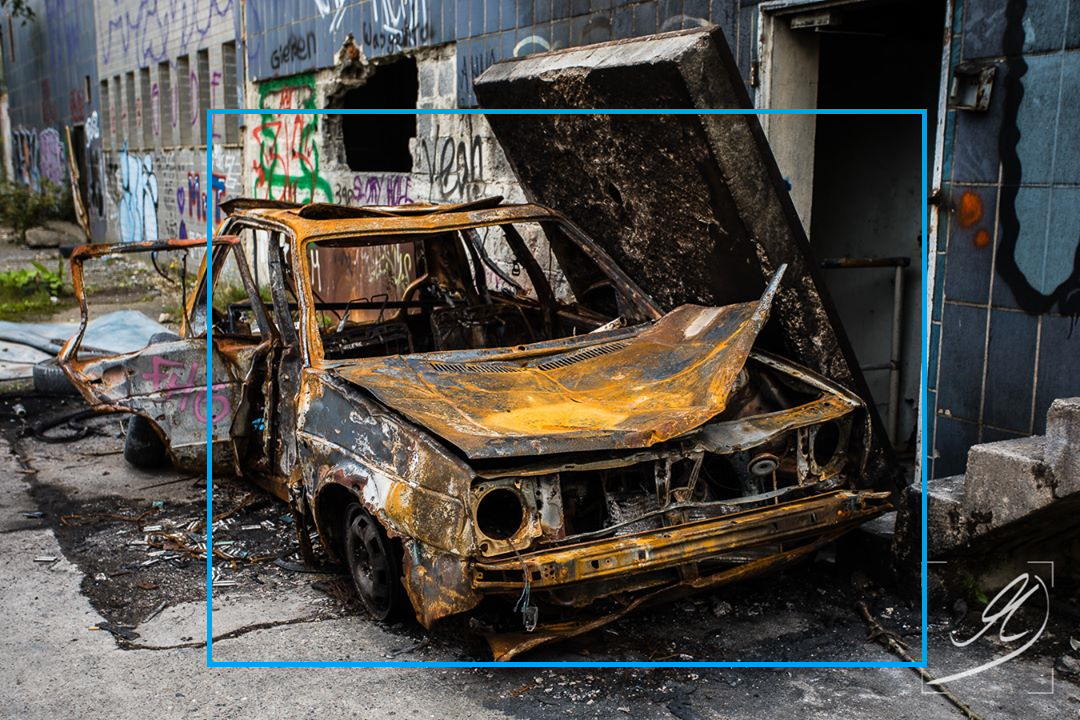}
  \label{fig:f2}}
  \caption{Bounding Box Annotations}
  \label{bbox_fire_infra}
\end{figure}

%nayem
%How we make the ground truth of the visual attention for each of the test data
%How a human do that independently
%Criteria how an human decide the attention location
To quantify the correctness of focusing attention by deep learning networks trained on our dataset, the test images are annotated by six human annotators. We do this to compare the human way of paying attention with the neural network. The annotation task has been performed following the standard annotations guidelines from the PASCAL Visual Object Classes(VOC) Challenge \cite{Everingham2010}. Each of our annotators has been trained before the task. Then, they have been asked to draw bounding boxes to the parts of the test images where their attention is intuitively drawn to infer the disaster class. A bounding box is drawn around the disaster region  making sure that all of the visible context of the disaster class are tightly inside the bounding box. These bounding box images are later compared with the attention-maps provided by CAM and TLAM  to calculate mean Intersection over Union (mIoU). Figure \ref{bbox_fire_infra} shows the examples of bounding boxes for images from Fire Disaster and Infrastructure Damage classes.
%and the ImageNet Challenge \cite{Su2012}.

\subsubsection{Visualizing  Attention and verification}
% To verify the correctness we write some code in which an image is read. A human annotator confirms multiple regions from an image in where the attentions must be given. Then a bounding box has drawn for each confirmed regions. When the selections are finalized we take the coordinates for each regions from the actual input. We set black color on every pixels from selected regions and make other pixels white. This process gives us a binary image as output. This is done for each test images by 3 annotators independently.
% For see the classifier's attention, we produce heatmaps for each of the test image. Based on a fixed threshold (0.6) over the heat map, we identify the region where the classifier provides most of the attention. Then we have created a mask where we put 1s where the classifier puts it's attention mostly. We put 0 on the other regions. 

% Then we calculate the mean intersection over Union between human annotator's attention and classifier's attention.

This experiment is performed to quantify the attention localization capability of classifiers trained on our dataset. 
% Annotators first put bounding boxes on the parts of test images where they think their attention is being drawn naturally to infer the disaster classes. The bounding boxes are used later to compare the attention provided by CAM and TLAM. %later transformed to masks by putting highest intensity value to the pixels bounded by the bounding box and zeros to rest of the pixels.

The CAM and TLAM output images are transformed to binarized masks by making normalized attention values greater than a threshold to have intensity value of one and rest of the pixels to zero. The thresholds are 0.15 and 0.10 for CAM and TLAM, respectively. We opted for a lower threshold in case of TLAM because attention heatmap outputs of TLAM are very fine and thin compared to CAM. The annotated test images are also binarized by making pixels in the bounding box to have intensity value of one and the rest to zero. After that, the amount of overlapping between the two masks are calculated using Intersection Over Union (IOU) method. This procedure is performed over all the test images. Then the mean IOU is calculated, which is the quantified score.

\subsubsection{Performance Measurement} We present our classification result in terms of  accuracy and macro F1-score. Moreover, we have calculated the mIoU to show how well the classifier's attention overlaps with the human attention. 

\section{Experimental Results and Discussions}
In this section we present the results of the experiments that we have designed in last section. We compare our results with UCI dataset to show the efficacy of our proposed dataset. 

\subsection{Diversity}
\begin{figure}[!ht]

 \centering
  \includegraphics[width = 9 cm]{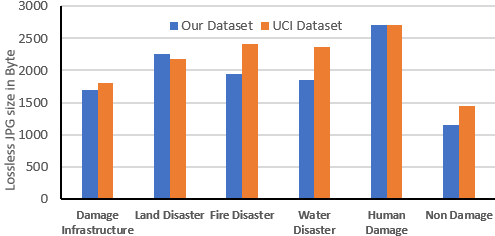}
  \caption{ Lossless JPG size in byte of our dataset vs UCI dataset }
  \label{fig:diversity}
\end{figure}

Figure \ref{fig:diversity} shows the lossless JPG file size of average image for each class of our dataset vs UCI dataset. It is observed that the average images for four out of six classes of our dataset have less byte size and thus contain more information. Therefore,  our dataset is more diverse than UCI dataset.

\subsection{Performance and Attention Analysis}
\subsubsection{Five-fold Cross Validation}
Table \ref{tab:crossval} shows the performance of the CAM and TLAM for the 5-fold cross validation experiment. We report accuracy and macro average F1 score for each of the fold tested.

\begin{table}[h]
\caption{Cross Validation Summary for CAM and TLAM} \label{tab:crossval}
\centering
\begin{tabular}{|l|c|c|c|c|}
\hline
\multirow{2}{*}{} & \multicolumn{2}{c|}{CAM} & \multicolumn{2}{c|}{TLAM} \\ \cline{2-5} 
 & Accuracy & \begin{tabular}[c]{@{}c@{}}F1 score\\ (Macro Avg)\end{tabular} & Accuracy & \begin{tabular}[c]{@{}c@{}}F1 score\\ (Macro Avg)\end{tabular} \\ \hline
Fold 1            & 0.96        & 0.89       & 0.96        & 0.89        \\ \hline
Fold 2            & 0.96        & 0.90       & 0.96        & 0.88        \\ \hline
Fold 3            & 0.95        & 0.89       & 0.96        & 0.88        \\ \hline
Fold 4            & 0.96        & 0.92       & 0.97        & 0.92        \\ \hline
Fold 5            & 0.96        & 0.90       & 0.96        & 0.88        \\ \hline
\end{tabular}
\end{table}

%While investigating the reason we found that in the fold4 test, the precision and recall for  Human Damage class are 0.92  and 0.55 that result poor f1 score for Human Damage. Almost half of  of the images with actually Human Damage class are misclassfied as Damaged Infrastructure and Non-Damage. Further investigation reveals that the train images that are selected from Human damage class in fold4 do not have clear visualization of features (blood, scratches, wounds) on the human body. 

It can be easily observed that the  accuracy for each of the fold is close to 0.96 for CAM. Also, the F1-score is around 0.90.  Similar results are also observed for TLAM. Accuracy scores are 0.96  and macro F1 scores are around 0.88 for all folds except fold 4. Fold 4 has slightly better performance (accuracy: 0.97 and macro F1 score 0.92). 5-fold cross validation results with both CAM and TLAM suggest that our proposed dataset is well structured and uniform throughout its extent. %No particular parts are biased for any class.

%As an example, we carefully see the confusion matrix for Fold 3 with CAM where the macro Precisionuracy and macro F1 score are slightly low (0.95 and 0.89, respective).  

% \begin{figure}[!h]
%  \centering
%  %\begin{minipage}[b]{0.52\textwidth}
%   \includegraphics[width= .50\textwidth]{crossValidationFold4_2.png}
%  %\end{minipage}%
% % \hspace{0.05cm}
%  \caption{Confusion Matrix for fold 4}
% \end{figure}

\subsubsection{Testing performance}
In table  \ref{tab:testCAMTLAM}, we report  the detailed performance result of test procedure. Discriminating features such as fire, smoke, flame help in better classification of fire images. Moreover, as the Water Disaster images of our dataset have unique characteristics, the F1 score is high for this class.

\begin{table*}[!ht]
%\label{testsummary}
\centering
\caption{Performance  Summary for CAM and TLAM on Test Data} \label{tab:testCAMTLAM}

\begin{tabular}{|c|c|c|c|c|c|c|c|c|}
\hline
\multicolumn{1}{|l|}{\multirow{2}{*}{}} &
  \multicolumn{2}{c|}{\begin{tabular}[c]{@{}c@{}}CAM \\ trained on\\ proposed training set\end{tabular}} &
  \multicolumn{2}{c|}{\begin{tabular}[c]{@{}c@{}}CAM\\ trained on\\ UCI dataset\end{tabular}} &
  \multicolumn{2}{c|}{\begin{tabular}[c]{@{}c@{}}TLAM\\ trained on\\ proposed training set\end{tabular}} &
  \multicolumn{2}{c|}{\begin{tabular}[c]{@{}c@{}}TLAM\\ trained on\\ UCI dataset\end{tabular}} \\ \cline{2-9} 
\multicolumn{1}{|l|}{} & Precision & F1   & Precision & F1   & Precision & F1   & Precision & F1   \\ \hline
Infrastructure Damage  & 0.91      & 0.93 & 0.71      & 0.81 & 0.87      & 0.92 & 0.67      & 0.78 \\ \hline
Fire Damage            & 1.00      & 0.98 & 0.03      & 0.03 & 1.00      & 1.00 & 0.00      & 0.00 \\ \hline
Human Damage           & 0.91      & 0.92 & 0.00      & 0.00 & 0.94      & 0.97 & 0.00      & 0.00 \\ \hline
Water Disaster         & 1.00      & 1.00 & 0.91      & 0.94 & 1.00      & 0.98 & 0.94      & 0.94 \\ \hline
Land Disaster          & 0.94      & 0.94 & 0.06      & 0.06 & 1.00      & 0.94 & 0.00      & 0.00 \\ \hline
Non Damage             & 0.97      & 0.96 & 1.00      & 0.65 & 1.00      & 0.99 & 0.70      & 0.61 \\ \hline
Macro Average          & 0.96      & 0.96 & 0.45      & 0.42 & 0.97      & 0.97 & 0.38      & 0.39 \\ \hline
\end{tabular}
\end{table*}

\begin{figure*}[!ht]

 \centering
  \includegraphics[width=0.9\textwidth]{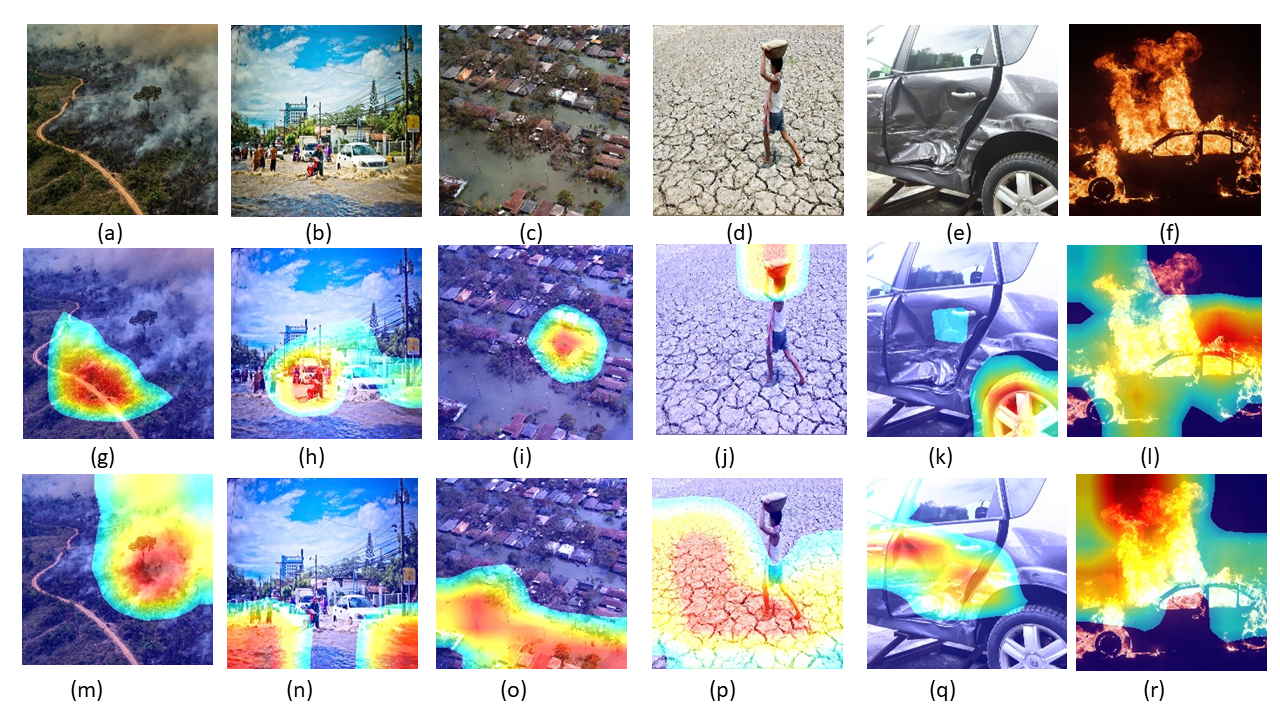}
 \caption{Top row: Input images from (a) Fire Disaster, (b) Water Disaster, (c) Water Disaster, (d) Land Disaster, (e) Infrastructure Damage, (f) Fire Disaster; Middle row: Wrong attention (Using CAM trained with UCI dataset); Last row: Correct attention (Using CAM trained with our dataset)}
 \label{fig:image6x3Intro}
\end{figure*}

%\vspace{-10 pt}

Table \ref{tab:testCAMTLAM} clearly shows the efficacy of our dataset in training the classifier models. The test images are classified with significantly high F1-score. The macro average F1 score for CAM and TLAM are: 0.96 and 0.97, respectively. In contrast the macro F1 scores of the same classifiers are very low: 0.42 and 0.39 when the classifiers are trained with UCI dataset. Most importantly, the UCI dataset cannot make the classifiers learn effective discriminative features for Human Damage, Fire Disaster, and Land Disaster. We observed that for most of the test images, the classifiers have put attentions completely in the wrong regions.

Figure \ref{fig:image6x3Intro} presents some samples of attention heatmaps. In this figure, first row has the input images, middle row contains examples of misplaced attention. Third row shows that attention is moved to the correct region when we train CAM with our dataset. The heatmap represents the attention intensity. Reddish heatmap indicates higher attention.

 \ref{fig:image6x3Intro}(a) is a test image from Fire Disaster class. But UCI trained CAM model has put it's attention on the road as seen in the image \ref{fig:image6x3Intro}(g). In contrast, CAM trained with our dataset learns to pay it's attention correctly on the smoke region as shown in \ref{fig:image6x3Intro}(m).  
 
  A test image from Water Disaster class is in \ref{fig:image6x3Intro}(b) and the attention heatmap is shown in \ref{fig:image6x3Intro}(h) when CAM is trained with UCI. It is easy to see that classifier has put it's attention on the vehicles and human resulting in wrong classification. The image \ref{fig:image6x3Intro}(n) shows that the attention is in the expected water region; thus the CAM model could correctly classify it as Water Disaster image. 
  
  Similarly, \ref{fig:image6x3Intro}(d) is a Draught image that falls under the Land Disaster class. UCI trained CAM pays attention towards the human in \ref{fig:image6x3Intro}(j) whereas the CAM model trained with our dataset puts attention on the dry and fractured land in \ref{fig:image6x3Intro}(j) to classify the image as Land Disaster.
 
 We have also devised an experiment to show how well our dataset generalizes on UCI dataset. As UCI dataset does not have any explicit test set, we randomly pick 40 images from each class and make a UCI test set of 240 images. Then we compare the classification performance on UCI test set using CAM model trained on both our proposed dataset and rest of the UCI dataset. To make the result unbiased, we perform the random picking and testing five times. The average classification accuracy of the five runs of testing on UCI test set is 71.25\%  with CAM trained on our dataset and 68.62\% with CAM trained on UCI dataset. So, our dataset generalizes well on UCI dataset.

%Therefore, the performance for the non-damage %class classification is the highest.% All other classes except damaged infrastructure got F1-score around 0.85. However, infrastructure damage has comparatively less F-score: 0.76.

\subsubsection{BOWFIRE}
To further show the generalization capability, we have evaluated our model on benchmark disaster dataset. We have used the Bowfire test set \cite {bowfire}  that has only two classes: Fire and No-Fire. We have tested the Bowfire test set with CAM and TLAM that are trained with the proposed dataset. We observe that the F1 score  of  fire class with CAM (trained with our dataset) is $0.84$, whereas, the F1 score  reported in the paper \cite{muhammad2018early} for \cite{bowfire} is in the range of $0.6-0.7$ and for \cite{ref23} it is in range $0.5-0.6$. When we have performed the experiment with TLAM (trained with our dataset), we have got exactly the same F1 score  $0.84$. %The improved F1 scores suggest that the models trained with proposed dataset learn the parameters more appropriately than those reported in the papers \cite{bowfire}\cite{ref23}.

\subsubsection{Classifier's Attention vs Human Attention}
In order to show the quality of our dataset, we calculate the mean Intersection over Union (mIoU) of human attention and classifier's attention for our test data while the classifier has been trained with our training set. Additionally, we calculate mIou on test set for classifier trained with UCI dataset. Table \ref{fig:mIoU} shows that the mIoU is significantly higher when CAM has been trained with our training set. The mIoU for CAM trained on our proposed dataset is $0.53$ whereas the mIoU drops to  $0.45$ for CAM trained with UCI dataset. After doing the same experiment with TLAM, the mIoU significantly drops from $0.31$ to $0.18$. 

To show how subcategorization empowers classifiers' attention, we have trained CAM with all images except the images from Wildfire subcategory. The test mIoU drops from 0.53 to 0.5. Furthermore, we again train a new CAM model discarding Wildfire and Drought images. The test mIoU then drops to 0.49. 

Apart from CAM and TLAM, we have also experimented with recent attention module GradCam++ \cite{chattopadhay2018grad}. We train using resnet-101 architecture on both our dataset and UCI dataset. We then compare the overlap agreement between human attention and classifier's attention. Our dataset yields an mIOu of 0.56 and for UCI it is 0.49. So, GradCam++ also shows the efficacy of our dataset.

The experimental results suggest that the structure of the proposed dataset is such well organized and diverse that attention localization capability of classifiers are much improved.
% the classifiers learn to pay their attentions towards the same input image locations as humans do. On the other hand, training with UCI is unable to teach the classifier to provide it's attention to the appropriate locations. 

\begin{table}[hbt!]

\centering
\caption{Overlap agreement between  human attention and classifier attention} \label{tab:attenionOverlap}
\label{fig:mIoU}
\begin{tabular}{|c|c|c|}
\hline
    &  \makecell{Human-Classifier \\ Attention Overlap \\ (Mean IoU)} &  \makecell{Human-Classifier \\Attention  Overlap \\  (Mean IoU)} \\
 \hline
 Training Set & \makecell{Proposed \\Dataset} &\makecell{UCI \\ Dataset}   \\
 \hline
%  Test Set & Our Test Set & Our Test Set \\
%  \hline
 CAM & 0.53 & 0.45 \\
 \hline
 TLAM & 0.31 & 0.18 \\
 \hline

\end{tabular}
\end{table}

\section{Conclusion}
Accumulating a comprehensive image dataset for disaster detection task is very challenging, especially, images that contain representative information for different classes. Also, it is a difficult task to provide annotation for such images. To reduce biases in the classification process we have provided class information for each image by three different individuals. Also six annotators provided bounding box annotations for test images. In future, we plan to provide bounding box ground truth annotations for all the images of our dataset. More challenging issue is to come up with a proper assessment mechanism for such datasets. It is often not enough to look only at the quantitative measures such as classification accuracy, precision, recall, F1 score, etc. In this work we try analysing the performance of the attention based classifiers not only to show that, systems trained with our dataset can outperform exact same systems trained with other dataset, moreover, we show that visually and numerically, human level attention can be achieved if attention based classifiers are trained with our dataset.               

\section{Acknowledgment}
This project is supported by ICT Division, Government of Bangladesh, and Independent University, Bangladesh (IUB).

\bibliographystyle{IEEEtran}
\bibliography{IEEEICPR2020}

\begin{thebibliography}{10}
\providecommand{\url}[1]{#1}
\csname url@samestyle\endcsname
\providecommand{\newblock}{\relax}
\providecommand{\bibinfo}[2]{#2}
\providecommand{\BIBentrySTDinterwordspacing}{\spaceskip=0pt\relax}
\providecommand{\BIBentryALTinterwordstretchfactor}{4}
\providecommand{\BIBentryALTinterwordspacing}{\spaceskip=\fontdimen2\font plus
\BIBentryALTinterwordstretchfactor\fontdimen3\font minus
  \fontdimen4\font\relax}
\providecommand{\BIBforeignlanguage}[2]{{%
\expandafter\ifx\csname l@#1\endcsname\relax
\typeout{** WARNING: IEEEtran.bst: No hyphenation pattern has been}%
\typeout{** loaded for the language `#1'. Using the pattern for}%
\typeout{** the default language instead.}%
\else
\language=\csname l@#1\endcsname
\fi
#2}}
\providecommand{\BIBdecl}{\relax}
\BIBdecl

\bibitem{OurGAIC2019paper}
Arif, A.~Omar, S.~Ashraf, A.~M. Rahman, M.~A. Amin, and A.~A. Ali, ``A
  comparative study on disaster detection from social media images using deep
  learning,'' in \emph{Global AI Congress}, 2019.

\bibitem{rizk2019computationally}
Y.~Rizk, H.~S. Jomaa, M.~Awad, and C.~Castillo, ``A computationally efficient
  multi-modal classification approach of disaster-related twitter images,'' in
  \emph{Proceedings of the 34th ACM/SIGAPP symposium on applied computing},
  2019, pp. 2050--2059.

\bibitem{panag2017people}
P.~Giannakeris, K.~Avgerinakis, A.~Karakostas, S.~Vrochidis, and
  I.~Kompatsiaris, ``People and vehicles in danger - a fire and flood detection
  system in social media,'' 06 2018.

\bibitem{Khan2018Early}
K.~Muhammad, J.~Ahmad, and S.~Baik, ``Early fire detection using convolutional
  neural networks during surveillance for effective disaster management,''
  \emph{Neurocomputing}, 12 2017.

\bibitem{ko2011modeling}
B.~C. Ko, S.~J. Ham, and J.~Y. Nam, ``Modeling and formalization of fuzzy
  finite automata for detection of irregular fire flames,'' \emph{IEEE
  Transactions on Circuits and Systems for Video Technology}, vol.~21, no.~12,
  pp. 1903--1912, 2011.

\bibitem{2nddataset}
S.~Verstockt, T.~Beji, P.~De~Potter, S.~Hoecke, B.~Sette, B.~Merci, and
  R.~Van~de Walle, ``Video driven fire spread forecasting (f) using multi-modal
  lwir and visual flame and smoke data,'' \emph{Pattern Recognition Letters},
  vol.~34, pp. 62 -- 69, 01 2013.

\bibitem{bowfire}
D.~Chino, L.~Avalhais, J.~Rodrigues~Jr, and A.~Traina, ``Bowfire: Detection of
  fire in still images by integrating pixel color and texture analysis,'' 08
  2015, pp. 95--102.

\bibitem{foggiadataset}
P.~{Foggia}, A.~{Saggese}, and M.~{Vento}, ``Real-time fire detection for
  video-surveillance applications using a combination of experts based on
  color, shape, and motion,'' \emph{IEEE Transactions on Circuits and Systems
  for Video Technology}, vol.~25, no.~9, pp. 1545--1556, 2015.

\bibitem{firoj2017image4act}
F.~Alam, M.~Imran, and F.~Ofli, ``Image4act: Online social media image
  processing for disaster response,'' 07 2017, pp. 601--604.

\bibitem{hussein2018multi}
H.~Mozannar, Y.~Rizk, and M.~Awad, ``Damage identification in social media
  posts using multimodal deep learning,'' 05 2018.

\bibitem{ourdataset}
\BIBentryALTinterwordspacing
Disaster dataset. [Online]. Available:
  \url{https://niloy193.github.io/Disaster-Dataset}
\BIBentrySTDinterwordspacing

\bibitem{alexnet}
A.~Krizhevsky, I.~Sutskever, and G.~E. Hinton, ``Imagenet classification with
  deep convolutional neural networks,'' in \emph{Advances in neural information
  processing systems}, 2012, pp. 1097--1105.

\bibitem{Avgerinakis2017VisualAT}
K.~Avgerinakis, A.~Moumtzidou, S.~Andreadis, E.~Michail, I.~Gialampoukidis,
  S.~Vrochidis, and Y.~Kompatsiaris, ``Visual and textual analysis of social
  media and satellite images for flood detection @ multimedia satellite task
  mediaeval 2017,'' in \emph{MediaEval}, 2017.

\bibitem{damageassess}
D.~T. Nguyen, F.~Ofli, M.~Imran, and P.~Mitra, ``Damage assessment from social
  media imagery data during disasters,'' in \emph{Proceedings of the 2017
  IEEE/ACM International Conference on Advances in Social Networks Analysis and
  Mining 2017}, 2017, pp. 569--576.

\bibitem{zhou2016learning}
B.~Zhou, A.~Khosla, A.~Lapedriza, A.~Oliva, and A.~Torralba, ``Learning deep
  features for discriminative localization,'' in \emph{Proceedings of the IEEE
  conference on computer vision and pattern recognition}, 2016, pp. 2921--2929.

\bibitem{jetley2018learn}
S.~Jetley, N.~A. Lord, N.~Lee, and P.~H. Torr, ``Learn to pay attention,''
  \emph{arXiv preprint arXiv:1804.02391}, 2018.

\bibitem{californiawildfireesquire}
\BIBentryALTinterwordspacing
J.~Holmes and K.~Sherin. (October 28, 2019) 40 photos show the incredible
  destruction wrought by the 2019 california wildfires. [Online; accessed
  23-January-2020]. [Online]. Available:
  \url{www.esquire.com/news-politics/g29610550/california-wildfire-photos-2019/?slide=9}
\BIBentrySTDinterwordspacing

\bibitem{brazilcbsnews}
\BIBentryALTinterwordspacing
Pictures from the amazon rainforest fires. [Online; accessed 25-January-2020].
  [Online]. Available:
  \url{https://www.cbsnews.com/pictures/pictures-amazon-rainforest-fires-in-brazil/17}
\BIBentrySTDinterwordspacing

\bibitem{portal:yemenwar}
\BIBentryALTinterwordspacing
{}. (August 18, 2018) Yemen conflict: Un experts detail possible war crimes by
  all parties. [Online; accessed 28-January-2020]. [Online]. Available:
  \url{https://www.bbc.com/news/world-middle-east-45329220}
\BIBentrySTDinterwordspacing

\bibitem{intelimage}
\BIBentryALTinterwordspacing
J.~B. Puneet~Bansal, ``Intel image classification,'' January 2019. [Online].
  Available: \url{https://www.kaggle.com/puneet6060/intel-image-classification}
\BIBentrySTDinterwordspacing

\bibitem{deng2009imagenet}
J.~Deng, W.~Dong, R.~Socher, L.-J. Li, K.~Li, and L.~Fei-Fei, ``Imagenet: A
  large-scale hierarchical image database,'' in \emph{2009 IEEE conference on
  computer vision and pattern recognition}.\hskip 1em plus 0.5em minus
  0.4em\relax Ieee, 2009, pp. 248--255.

\bibitem{Everingham2010}
M.~Everingham, L.~{Van Gool}, C.~K. Williams, J.~Winn, and A.~Zisserman, ``{The
  pascal visual object classes (VOC) challenge},'' \emph{International Journal
  of Computer Vision}, vol.~88, no.~2, pp. 303--338, 2010.

\bibitem{muhammad2018early}
K.~Muhammad, J.~Ahmad, and S.~W. Baik, ``Early fire detection using
  convolutional neural networks during surveillance for effective disaster
  management,'' \emph{Neurocomputing}, vol. 288, pp. 30--42, 2018.

\bibitem{ref23}
T.~Celik and H.~Demirel, ``Fire detection in video sequences using a generic
  color model,'' \emph{Fire Safety Journal}, vol.~44, 2009.

\bibitem{chattopadhay2018grad}
A.~Chattopadhay, A.~Sarkar, P.~Howlader, and V.~N. Balasubramanian,
  ``Grad-cam++: Generalized gradient-based visual explanations for deep
  convolutional networks,'' in \emph{2018 IEEE Winter Conference on
  Applications of Computer Vision (WACV)}.\hskip 1em plus 0.5em minus
  0.4em\relax IEEE, 2018, pp. 839--847.

\end{thebibliography}

\end{document}